\documentclass[letterpaper, 10 pt, conference]{ieeeconf}  

\IEEEoverridecommandlockouts                              

\usepackage{graphics} 
\usepackage{graphicx}
\usepackage{amsmath} 
\usepackage{amssymb}  
\usepackage{mathtools}
\usepackage{hyperref}

\usepackage{xcolor}

\usepackage[symbol]{footmisc}

\title{\LARGE \bf
Learning Bipedal Walking On Planned Footsteps \protect\\
For Humanoid Robots
}

\author{Rohan P. Singh$^{1,2}$, Mehdi Benallegue$^{1}$, Mitsuharu Morisawa$^{1}$, \\
Rafael Cisneros$^{1}$, Fumio Kanehiro$^{1,2}$
\thanks{$^{1}$CNRS-AIST JRL (Joint Robotics Laboratory) IRL, National Institute of Advanced Industrial Science and Technology (AIST), Japan.
$^{2}$ University of Tsukuba, Ibaraki, Japan.
        {\tt\small rohan-singh@aist.go.jp}}%
}

\begin{document}

\maketitle
\thispagestyle{empty}
\pagestyle{empty}

\begin{abstract}

Deep reinforcement learning (RL) based controllers for legged robots have demonstrated
impressive robustness for walking in different environments for several robot platforms.
To enable the application of RL policies for humanoid robots in real-world
settings, it is crucial to build a system that can achieve robust walking in any direction,
on 2D and 3D terrains, and be controllable by a user-command.
In this paper, we tackle this problem by learning a policy to follow a given step sequence.
The policy is trained with the help of a set of procedurally generated step sequences
(also called footstep plans).
We show that simply feeding the upcoming 2 steps to the policy is sufficient to
achieve omnidirectional walking, turning in place, standing, and climbing stairs.
Our method employs curriculum learning on the complexity of terrains,
and circumvents the need for reference motions or pre-trained weights.
We demonstrate the application of our proposed method to learn RL policies for 
2 new robot platforms - HRP5P and JVRC-1 - in the MuJoCo
simulation environment. The code for training and evaluation is available online.
\footnote[2]{\url{https://github.com/rohanpsingh/LearningHumanoidWalking}}.
\end{abstract}

\section{Introduction}

Learning-based methods such as model-free, deep reinforcement learning (RL) for control
have shown us a new direction for controlling legged robots, in the recent few years.
RL policies can be trained for balancing tasks, locomotion tasks, and a wide range of
complex manipulation skills. While several works have shown impressive results demonstrating
bipedal walking (in simulation and on real robots), wider adoption of such controllers
for practical applications still lies ahead in the future.

An important aspect of practical robots is the ability of the underlying controller
to track a user-specified command - which may describe the desired mode of walking.
More concretely, for real-world deployment, it is useful to realize a controller which is
able to execute walking on curved paths, on flat-terrain and on stairs, forward and
backwards walking, and be able to stand still, in a user-commanded fashion.
The bipedal robot should also be able to easily transition between these different modes,
ideally without switching to a different controller. Traditionally, frameworks that are based
on a model-based controls achieve this through a footstep plan consisting of target feet positions
and orientations, combined with a finite-state machine (FSM). Footstep plans also greatly reduce
the uncertainty in the behavior of the controller as they allow us to know, roughly, in advance
when and where the robot is going to place its feet. This improves the overall safety of a
demonstration.

\begin{figure}[t]
  \vspace{-0.5cm}\includegraphics[width=\linewidth]{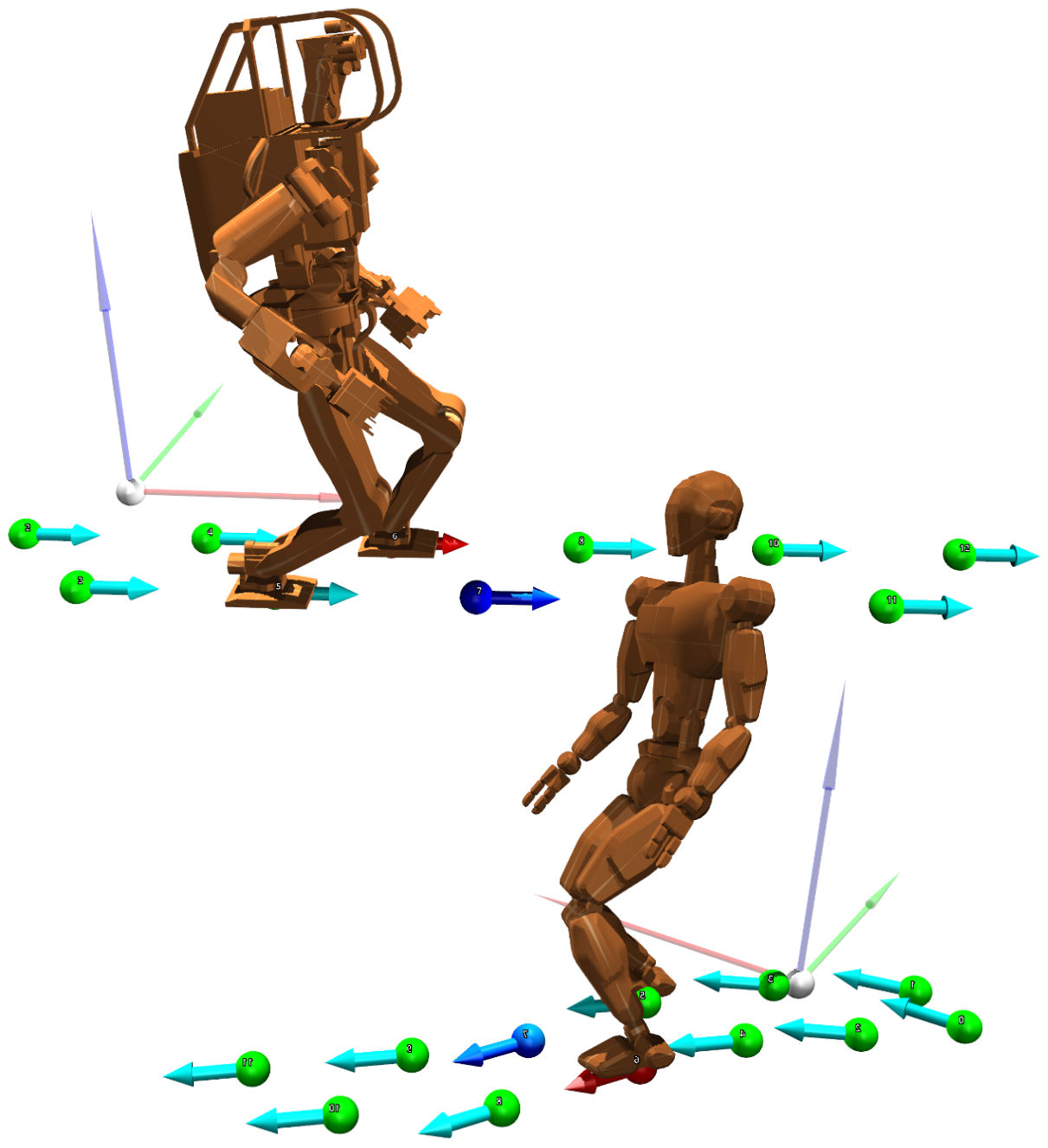}
  \caption{\textbf{Dynamic simulation} of HRP-5P (walking on straight path) and JVRC-1 (walking on curved path).
  The robot's walking pattern can be controlled by easily
  modulating the upcoming 2 target steps, $\textcolor{red}{\mathbf{T_1}}$ and $\textcolor{blue}{\mathbf{T_2}}$.}
  \label{figure:hrp5p_mujoco}
  \vspace{-0.25cm}
\end{figure}

Is it possible to achieve such a behavior for a legged robot using Reinforcement Learning?

In this work, we focus precisely on this problem. We develop a novel technique for
the design of RL bipedal locomotion policies for human-scale humanoid robots
(vastly different from the more popular \textit{Humanoid} character in RL literature),
capable of omnidirectional walking on a flat floor and 3D terrain. The user-command
to the RL policy is provided in terms of the desired feet positions and root heading at the future
2 footsteps, from the robot's current state. Using our proposed method, we demonstrate a
\textit{single} policy that can achieve the following modes by merely manipulating the
2-steps input signal:

\begin{itemize}
    \item Forwards or backward walking on a straight path,
    \item Walking on curved path,
    \item Lateral walking (sidestepping),
    \item Ascending and descending stairs,
    \item Turning in place, and
    \item Quiet standing.
\end{itemize}

Specifically, we make the following contributions through this research: 

\begin{enumerate}
    \item We show, for the first time, the development of an RL policy for the
    locomotion of human-sized humanoid robots without the use of reference motions or demonstrations.
    We show applications on 2 humanoids - HRP-5P \cite{kaneko2019humanoid} and JVRC-1 \cite{jvrcpaper}
    within a realistic simulation environment.
    
    \item We show the use of \textit{oriented} stepping targets for achieving omnidirectional locomotion
    on 2D terrains and 3D stairs in conjunction with existing footstep planners, and evaluate the
    robustness of the developed policy (in the context of real robot application).
    
    \item We release the training and evaluation code along with the model files for JVRC-1 publicly to
    facilitate further research in this direction.
\end{enumerate}

\section{Related Work}

\textbf{RL for legged locomotion.}
There exist several works that show promising results for using reinforcement learning
for legged locomotion tasks on several robot platforms \cite{peng2017deeploco,liu2019implementation}
either on the real physical hardware or on a realistic, simulated version of them.
Hwangbo et. al. \cite{hwangbo2019learning} demonstrated a method for training RL policies in 
simulation and transferring to the real ANYmal robot - a medium-dog-sized quadruped platform. Natural-looking
and energy-efficient motion was developed by gradually changing the reward function during training. This is
one way of implementing \textit{curriculum learning}, and has been extensively used by later works to similar
effect.

Recent works on the Cassie robot have been presented that show robust RL policies for the real hardware in
outdoor environments. To achieve a periodic bipedal gait on Cassie, with swing and stance phases for each foot,
reward function specification based on a clock signal is used in \cite{siekmann2021sim}. In \cite{li2021reinforcement},
a parameterized gait is achieved with the use of reference motions based on Hybrid Zero Dynamics (HZD).


Traversing stair-like terrain relying solely on proprioceptive measurements has also been demonstrated
as a means of effectively overcoming the problem of accurate terrain estimation \cite{siekmann2021blind}.
Achieving robust blind locomotion means that the robot will recover from missteps if it experiences an
unexpected contact with the environment. In contrast, our focus in this work is to develop controllers that
actively use the provided terrain data in the form of desired footsteps.

\textbf{RL for humanoids.}
Yang et. al. \cite{yang2020learning} have presented results on the Valkyrie humanoid robot using a realistic
robot model for learning walking gaits, albeit only within the simulation environment. They rely on expert
demonstrations and a tracking reward to achieve a seemingly natural bipedal walk. It is important to note that
deploying RL policies on the physical hardware for a large humanoid robot such as Valkyrie and the HRP-series robots
is considerably more problematic than in the case of lighter bipedal robots (say, Cassie). Large and
heavy humanoid robots have strong and heavy legs (to support a heavy upper-body) and consequently, a 
high gear-ratio transmission systems with low backdrivability. Safety risks are also heightened in
the case of a bigger and more powerful robot.

The method proposed in DeepWalk \cite{rodriguez2021deepwalk} shows a single learned policy for a real
humanoid robot that can achieve omnidirectional walking on a flat floor. However, it is not clear from
whether the robot can attain a dynamically stable walk or not. The robot platform also appears to have
relatively large feet which may help the robot to avoid falls even under a
fragile RL policy. 

Results on the Digit robot shown in \cite{castillo2021robust} demonstrate a hierarchical framework comprising
of a high-level RL policy with low-level model-based regulators and balancing controllers.

Our work in this paper is more similar to ALLSTEPS \cite{xie2020allsteps} (and a very recent work
\cite{duan2022sim}) which proposed a curriculum-driven learning of stepping stones skills for the
Cassie robot. In contrast to \cite{xie2020allsteps} and \cite{duan2022sim}, our framework introduces
the use of ``oriented" stepping targets allowing the RL policy to consume the planned footsteps
from an existing planner, which typically contains orientation targets in addition to planned footstep
locations.

Unlike Cassie in ALLSTEPS, our reward design ensures that a reference motion is not required to bootstrap
the RL policy. Additionally, we analyze the performance of our policy under noisy state measurement -
which is of paramount importance for robotics application. We also discuss the differences in learned
behavior between Cassie and the humanoids robots in our experiments.

\begin{figure*}
  \includegraphics[width=\textwidth]{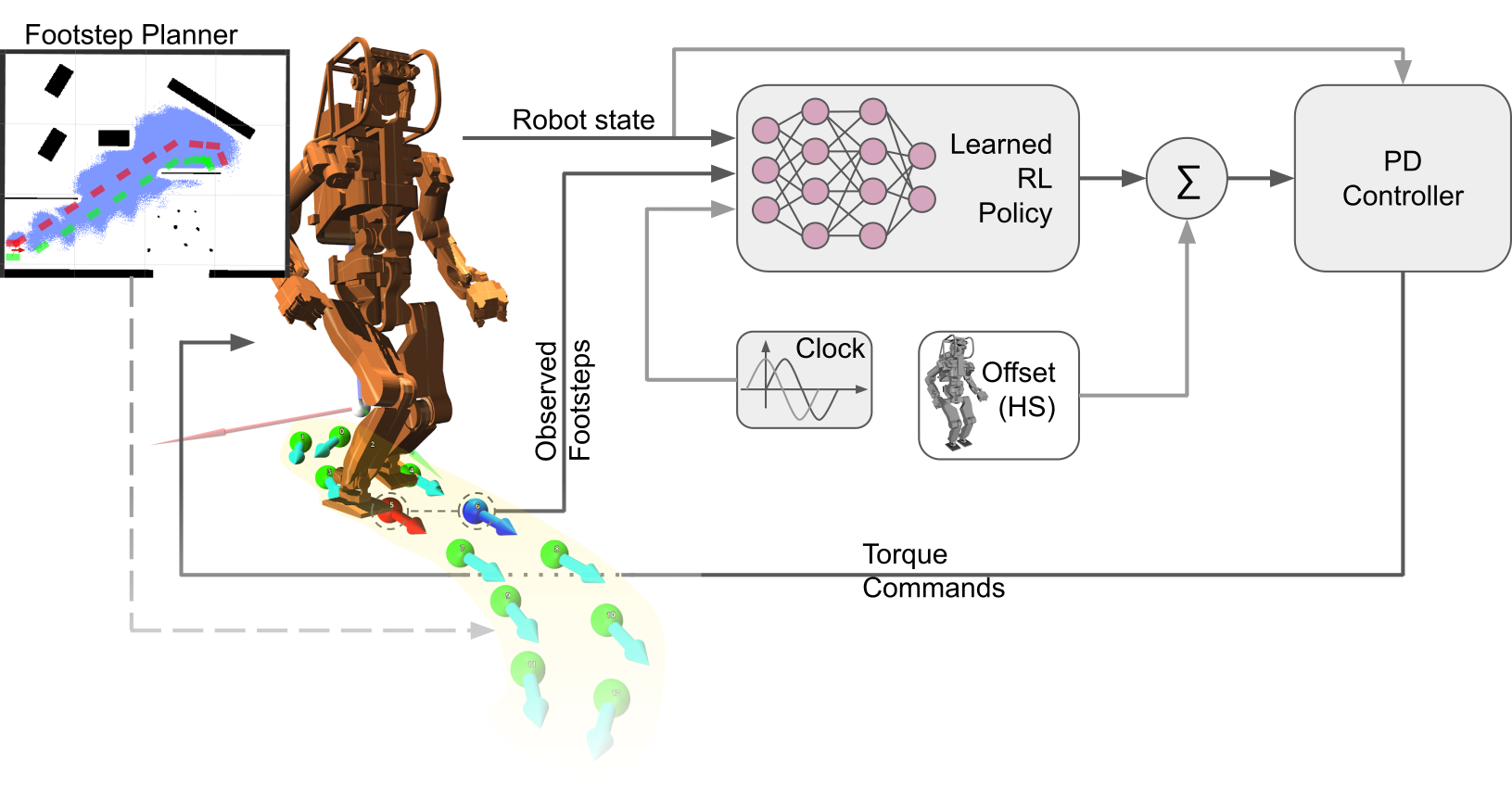}
  \vspace{-2.5cm}\caption{\textbf{Proposed hierarchical control structure.} The high-level controller
  is represented by the learned RL policy.
  The predictions made by RL policy $\pi$ are added to the neutral motor-positions, corresponding to
  the ``half-sitting" (HS) posture of HRP-5P, and then sent to the PD control loop. The input
  to the policy are the next two planned footsteps ($\textcolor{red}{\mathbf{T_1}}$ and
  $\textcolor{blue}{\mathbf{T_2}}$), the clock signal, and the robot
  state. Footstep planning is performed by conventional methods.}
  \label{figure:control_fw}
\end{figure*}



\section{Background}

\subsection{Robot Platforms}
HRP-5P is a high-powered, electrically-actuated, 53 degrees of freedom (DoF) humanoid robot
weighing over 100kg, standing tall at $182cm$ \cite{kaneko2019humanoid}. We freeze all the joints
corresponding to the upper body, arms, and hands and only actuate the leg joints - hip yaw, hip
roll, hip pitch, knee, ankle pitch, ankle roll. Thus, a total of 12 joints are controller by the RL policy.

JVRC-1 is a virtual robot model developed for the Japan Virtual Robotics Challenge \cite{jvrcpaper} with a
total weight of 62kg and $172cm$ height. Again, we freeze all joints other than the following joints - 
hip pitch, hip roll, hip yaw, knee, ankle roll, ankle pitch - in each leg. 

Description files for both robots were prepared for MuJoCo \cite{todorov2012mujoco}
from the original VRML models (we publicly release the MJCF model for JVRC-1 with this work).



\subsection{Reinforcement Learning}
RL concerns with the task of learning
to prescribe an action \textit{a} given an input \textit{s} with the aim of maximizing a reward
\textit{r}. The world is generally represented as a discrete-time Markov Decision Process
constituted by a continuous state space $\mathbf{S} \in \mathbb{R}^n$, an action space
$\mathbf{A} \in \mathbb{R}^m$, a state-transition function $p(s, a, s')$ and a reward function
$r(s,t)$. The state-transition function $p: S \times A \times S \rightarrow \begin{bmatrix} 0, 1 \end{bmatrix}$
defines the dynamics of the world and gives the probability density over the next state $s'$
when taking action $a$ in the current state $s$. $p$ is assumed to be unknown \textit{a priori}.
The reward function $r: S \times A \rightarrow \mathbb{R} $ provides a time-dependent scalar
signal at each state transition.

The policy $\pi(a|s)$ is defined as a stochastic mapping from states to action. The
goal in RL is to find a $\pi$ that maximizes the agent's expected T-horizon 
discounted return given by
$J(\pi_\theta) = \mathbb{E} \begin{bmatrix} \sum_{t=0}^{T} \gamma^t r(s_t, a_t) \end{bmatrix}$,
where $\gamma \in (0,1] $ is the discount factor. In the case of large, continuous state spaces we use a
parametric policy $\pi_{\theta}$ with parameters $\theta$ often representing the set of parameters
of a multi-layered perceptron (MLP). The policy is improved iteratively, by estimating the
gradient of $J(\pi_{\theta})$  and updating the policy parameters by performing
stochastic gradient ascent with a step size $\alpha$:
$\theta_{k+1} = \theta_{k} + \alpha \nabla J(\pi_{\theta_k}) $.
$\nabla J(\pi_{\theta_k})$ can be estimated with the experience collected from
trajectories sampled by following the policy $\pi_{\theta_k}$.

In this paper, we use the Proximal Policy Optimization (PPO) \cite{schulman2017proximal} algorithm which builds upon
the vanilla policy gradient method described above, to increase sample efficiency while avoiding policy collapse.


\section{Methods}

\subsection{Control Structure}
The hierarchical control framework adopted by us consists of a higher-level RL policy that
makes joint position predictions at a slow update rate of 40Hz, and a lower-level PD controller
working at 1000Hz responsible for converting the desired joint positions to desired joint
torques. The PD controller uses relatively very low gains and consequently
the joint position tracking error can be significantly large. The policy learns to incorporate
this behavior of the PD loop into its predictions, and in fact, actively uses the tracking
error to generate interaction forces \cite{hwangbo2019learning}.
We expect the controller to work in association with a planner responsible for feeding desired footsteps
and desired root heading to the RL policy. Such a planner would ideally rely on effective environment perception,
to dynamically plan a path for the robot to follow.
The control structure overview is shown in Fig. \ref{figure:control_fw}.

\subsection{Policy Parameterization}

\textbf{Observation Space.} The input to our control policy comprises of the robot state, the external
state, and the clock signal. The robot state consists of the joint positions and joint velocities
of each actuated joint (only in the legs), roll and pitch orientation and angular velocity of the root (pelvis). 
The robot state is invariant to the yaw of the root in the world frame. This is fine because the policy
does not need to know its global yaw for stepping on the target steps, which are always expressed relative
to the robot. Naturally, it would be vital for the external footstep planner, responsible for
computing the relative footsteps, to have a full 6D estimate of the robot in the world frame.

The robot state vector is concatenated with the 8D external state vector and a 2D clock signal. The external
state is described by the 3D position and 1D heading of the two upcoming steps $\mathbf{T_1} = [ x_1, y_1, z_1, \theta_1 ]$
and $\mathbf{T_2} = [ x_2, y_2, z_2, \theta_2 ]$, defined in the frame of the robot's root as $\prescript{r}{}{\textbf{T}_1}$
and $\prescript{r}{}{\textbf{T}_2}$, respectively. The heading $\theta$ acts as a reference for the desired root
orientation of the robot. We use an observation of two steps as observing only the next one step may cause performance
degradation; while more steps may not give additional value \cite{xie2020allsteps, coros2008synthesis}.

A clock signal is needed due to the inclusion of the periodic reward terms.
Even though the clock signal can be represented by a single scalar for a cyclic phase variable
$\phi$ that increments from 0 to 1 at each timestep, we do a bijective projection of $\phi$ to a
2D unit cycle as follows: 

\begin{align}
\label{eq:clock}
\text{Clock} = \left\{ \sin \left( \frac{2\pi\phi}{L} \right), \cos \left( \frac{2\pi\phi}{L} \right)  \right\},
\end{align}

\noindent
where $L$ is the cycle period after which $\phi$ resets to 0. This projection is done to prevent an
abrupt jump of the clock input from 1 to 0 at the end of each cycle. Such an abrupt change may
lead to a non-smooth learning behavior \cite{yang2020learning}.

Our experiments show that this state vector is sufficient for the task.
This is because other seemingly important state features, such as the root height, 
are derivable from the sensor measurements of joint encoders and IMU,
and hence, are implicitly encoded into the robot state vector. Even the absence of
force-torque sensor readings is overcome thanks to the joint tracking error.

\textbf{Action Space.} The output of the policy is comprised of the desired joint positions
of the actuated joints in the robot's legs (12 for each robot). The predictions of the desired joint
positions from the network are added to fixed motor offsets corresponding to the robot's
half-sitting posture, before being sent to the lower-level PD controllers.



\subsection{Reward Design}
We explain each term of the full reward function in details.

\textbf{Bipedal Walking.}
A symmetrical bipedal walking gait is characterized by a periodic motion of the legs, alternating
between double-support (DS) phases, where both feet are in contact with the ground, and the single-
support (SS) phases, where one foot is swinging while the other supports the weight of the robot.

\begin{figure*}
  \includegraphics[width=\textwidth]{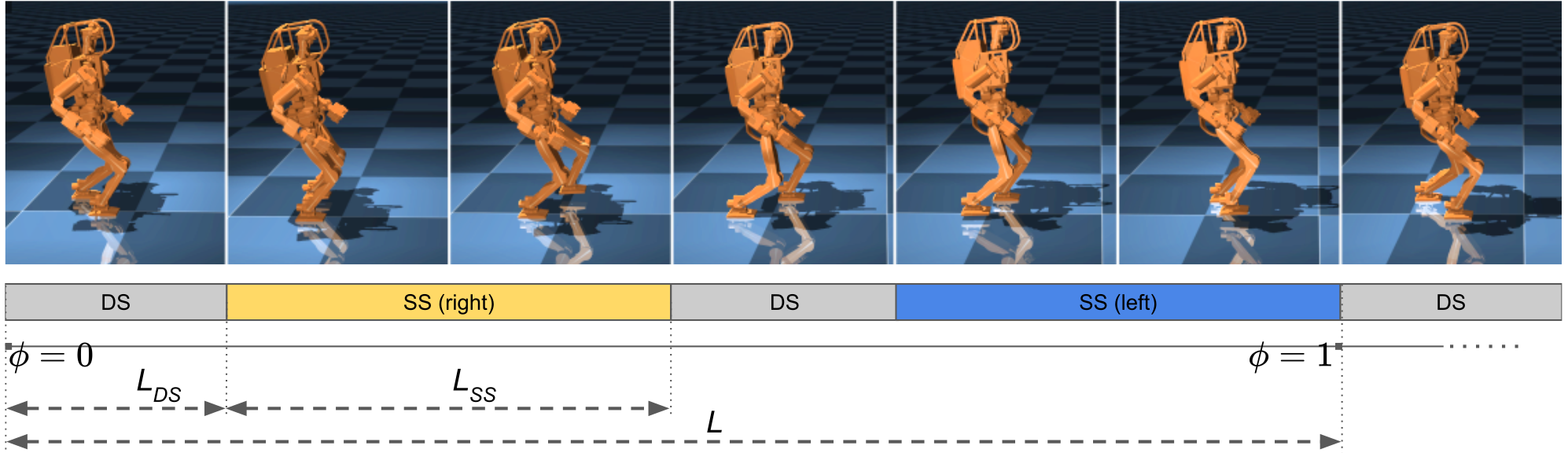}
  \caption{\textbf{Trained RL policy exhibits symmetric walking gait}. Screengrabs from left to right 
  are taken at 0\%, 15\%, 32\%, 50\%, 62.5\%, 78\% and 100\% of the gait cycle respectively. $L_{DS}$ is
  the double-support duration, and $L_{SS}$ is the single-support duration.  $L = 2\times(L_{DS} + L_{SS})$
  is the total cycle length.}
  \label{figure:walking_results}
\end{figure*}

To encode this pattern into the robot's behavior, we adopt the idea of periodic reward composition
proposed in \cite{siekmann2021sim}. Our main idea is to split one gait cycle into two SS phases 
(one for each foot) and two DS phases of fixed durations (refer to Fig. \ref{figure:walking_results}).
During a single-support phase, one foot is expected to make a static contact with the ground
while the other foot is expected to be swinging in the air. In the following SS phase,
the roles of the feet are swapped, and the foot previously on the ground is now expected to be
swinging and the previously swinging foot is expected to make the supportive contact with the
ground.

\begin{figure}
  \includegraphics[width=\linewidth]{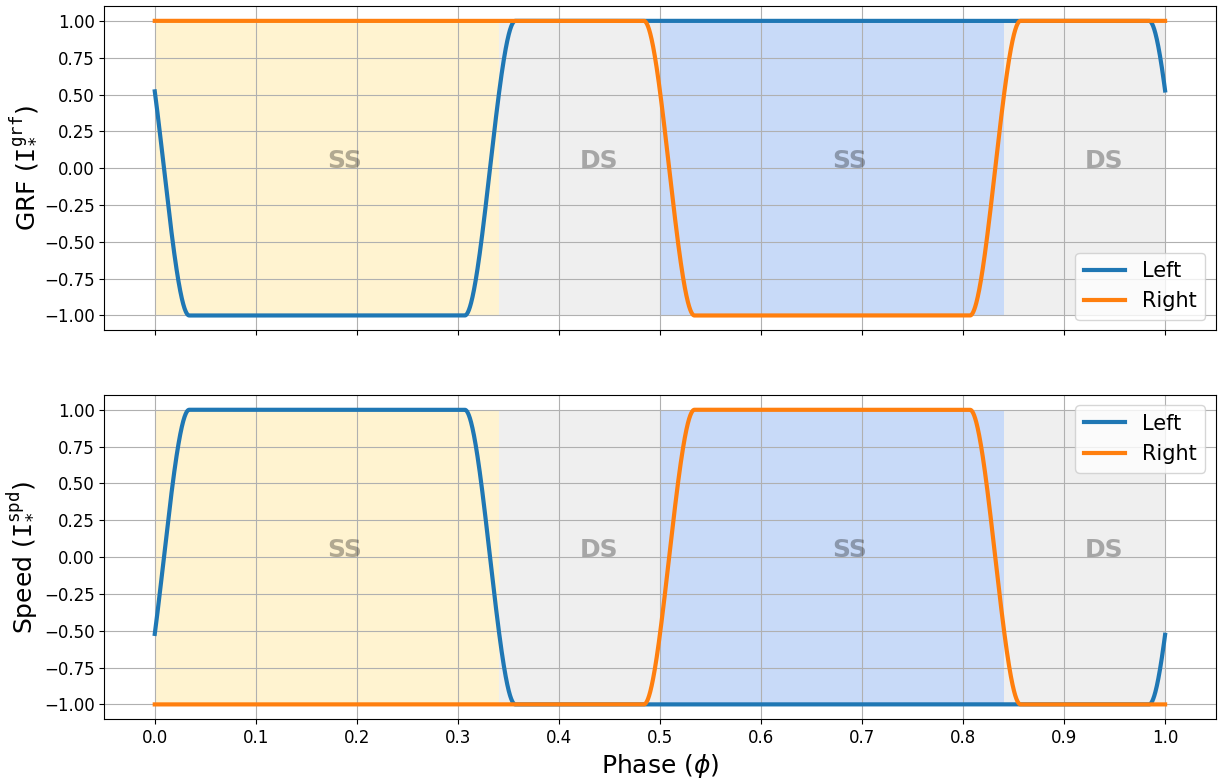}
  \caption{Plots of $I_{left}^{grf}$, $I_{right}^{grf}$, $I_{left}^{spd}$ and $I_{right}^{spd}$
  with respect to the phase variable $\phi$. The transitions at the phase boundaries are smoothened
  to encourage more stable learning \cite{siekmann2021sim}.}
  \label{figure:phases}
\end{figure}

We can implement the above behaviour with the help of ``phase indicator" functions
$I_{left}^{grf}(\phi),  I_{right}^{grf}(\phi)$ for regulating ground reaction forces
and $I_{left}^{spd}(\phi), I_{right}^{spd}(\phi)$ for regulating the speeds of the left
and right foot, respectively (see Fig. \ref{figure:phases}). During the SS phase,
the functions $I_{\operatorname{*}}(\phi) \in [-1, 1]$ incentivize higher speeds of
the swing foot while penalizing ground reaction forces (GRF) on the same foot, and
simultaneously penalize the speed and incentivize ground reaction forces on the support
foot. During the DS phase, speeds of both feet are penalized while the GRF are 
incentivized. We set the SS duration to 0.75s for HRP-5P and 0.80s for JVRC-1
and the DS duration to 0.35s for HRP-5P and 0.20s for JVRC-1.

The reward term for regulating the ground reaction forces at the feet $r_{\mathrm{grf}}$
and the term for the feet body speeds $r_{\mathrm{spd}}$ are computed as follows:

\begin{align}
& \text{$r_{\mathrm{grf}}$} = I_{left}^{grf}(\phi) \cdot F_{left} + I_{right}^{grf}(\phi) \cdot F_{right}
\label{eq:grf_reward} \\
& \text{$r_{\mathrm{spd}}$} = I_{left}^{spd}(\phi) \cdot S_{left} + I_{right}^{spd}(\phi) \cdot S_{right}
\label{eq:spd_reward}
\end{align}

From Eq. (\ref{eq:grf_reward}) and (\ref{eq:spd_reward}), and Fig. \ref{figure:phases}, it is easy
to see how the quantities (the normalized GRF $F_{left}$ and $F_{right}$ and the normalized body 
speeds $S_{left}$ and $S_{right}$) are rewarded positively or negatively according to $\phi$.
For example, when $\phi$ lies in the first single-support region of the gait cycle,
$I_{left}^{grf}(\phi)$ if close to -1 while $I_{right}^{grf}(\phi)$ is close to 1, meaning
that larger values of $F_{left}$ are rewarded negatively while larger values of $F_{right}$
lead to positive reward; that is, the left foot is swinging and the right foot creates the support.

For quiet standing, the DS phase is expanded to span the entire gait cycle. The quiet standing
mode is triggered when the observed two steps are zeroed, i.e., $\mathbf{T_1} = \mathbf{T_2} = \mathbf{0}$.

\textbf{Step Reward.}
In additional to the above periodic reward terms, we need to incentivize the robot to step
and orient the body according to the desired targets. The step reward is a combination of
two terms: the hit reward and the progress reward. 
The hit reward promotes the robot to place \textit{any} of its feet on the upcoming target
point. It is received only when either or both of the feet are within a distance \textit{target
radius} of step $\mathbf{T_1}$. The progress reward is responsible for encouraging the policy to
move the floating-base (\textit{root} body) towards the next target position from the current
position (both projected on the 2D plane). 

If the distance between the step $\mathbf{T_1}$ and the nearest foot is given by $d_{foot}$, 
while the distance between $\mathbf{T_1}$ and the root link is given by $d_{root}$, the step
reward is a weighted sum of the two terms:

\begin{multline} \label{eq:step_reward}
r_{\mathrm{step}} = k_{hit} \cdot \exp(-d_{foot}/0.25) + \\
          (1-k_{hit}) \cdot \exp(-d_{root}/2),
\end{multline} 

\noindent
where, coefficient $k_{hit} = 0.8$ is a tunable hyperparameter. 

Further, the root orientation term encourages the root orientation quaternion $q$ to be close
to the desired quaternion $\hat{q}$, obtained from the desired Euler angles of 0-degree roll and pitch, 
and a yaw equal $\theta_1$.

\begin{equation}
\label{eq:orient_reward}
r_{\mathrm{orient}} = \exp(-10 \cdot (1 - \langle q, \hat{q} \rangle^2)),
\end{equation}

\noindent
where $ \langle a,b\rangle $ denotes the inner product between $a$ and $b$.

We found that with the introduction of the progress reward, we do not need to explicitly
reward the robot to maintain a desired root velocity. Instead, by experiencing the curriculum of 
footsteps during training, the robot \textit{learns} to self-regulate its root velocity.

\textbf{Energy efficiency and other terms.}
Besides the above, we also desire several other characteristics of
the robot's gait. Specifically, we need to maintain the root height $h_{root}$ at a desired value $\hat{h}_{root}$,
equal to the height at the nominal posture.
This term is computed as follows:

\begin{align}
\label{eq:height_reward}
 r_{\mathrm{height}} = \exp(-40\cdot(h_{root} - \hat{h}_{root})^2).
\end{align}

We also introduce a term for encouraging the robot to maintain an upright posture, by introducing a
reward on the distance between the floor projection of the head positions $\prescript{x,y}{}{p_{head}}$ to the
root $\prescript{x,y}{}{p_{root}}$. This prevents the robot from developing a leaned-back behavior and
excessively swaying the upper body:

\begin{align}
    r_{\mathrm{upper}} = \exp (-10 \cdot \|\prescript{x,y}{}{p_{head}} - \prescript{x,y}{}{p_{root}} \| ^2).
    \label{eq:upper_reward}
\end{align}

Next, we introduce penalties for applied torque and action to increase energy efficiency of
the robot during locomotion and to prevent artefacts from developing. $r_{\mathrm{torque}}$ and $r_{\mathrm{action}}$
incentivize the joint torques $\tau$ and the predicted actions $a$ at the current timestep to be closer to
their respective values at the previous timestep, $\tau_{prev}$ and $a_{prev}$ respectively:

\begin{align}
    & r_{\mathrm{action}} = \exp (-5 \cdot \sum | a - a_{prev} | / 12), \\
    & r_{\mathrm{torque}} = \exp (-0.25 \cdot \sum | \tau - \tau_{prev} | / 12).
\end{align}

We do not include terms to regulate orientation of the feet. Finally, the full reward 
function is given by:

\begin{multline}
\label{eq:reward}
r = w_1 r_{\mathrm{grf}} + w_2 r_{\mathrm{spd}} + w_3 r_{\mathrm{step}} + \ldots \\
    w_4 r_{\mathrm{orient}} + w_5 r_{\mathrm{height}} + w_6 r_{\mathrm{upper}} +\ldots \\
    w_7 r_{\mathrm{action}} + w_8 r_{\mathrm{torque}},
\end{multline}

\noindent
where, the weights $w_1$, $w_2$, $w_3$, $w_4$, $w_5$, $w_6$, $w_7$, $w_8$ are set to
0.15, 0.15, 0.45, 0.05, 0.05, 0.05, 0.05, 0.05, respectively.
We did not aggressively fine-tune the reward weights, and use the same set for both robots.

\subsection{Early Termination and Initialization Conditions}
Besides the reward function, the initialization and termination conditions of the rollouts
have a strong impact on the learned behavior, as they can be used to prevent biases in the data
distribution towards unfavourable samples. As expressed in \cite{peng2018deepmimic}, this is
analogous to class imbalance related issues in supervised learning.

Inspired by previous works, we initialize each robot at its nominal posture: at which the robot can remain upright and prevent falling 
under the absence of external disturbances. For humanoids, this posture is commonly referred to as the 
``half-sitting" posture. We add a small amount of noise to the joint positions to let the policy experience
varying initial robot states; this will also be helpful for future deployment on the real hardware.
Since the global yaw angle of the robot's root is not observed by the policy, it is not important to randomize
the initial heading of the robot. Initial value of the phase variable $\phi$ can be $0$ or $0.5$.

Early termination conditions are needed to be set carefully such that the states where balance is
irrecoverable are avoided. We enforce two commonly used conditions - a fall conditions, reached when the root height
from the lowest point of foot-floor contact is less than a threshold, and a self-collision condition.
The height threshold is set as $60cm$ for both humanoids.
An episode ends either after a fixed number of control timesteps or when a termination 
condition is met.

\subsection{Footstep Generation}
Under our framework, a footstep is comprised of a 3D point with a heading vector attached to it
(represented by a scalar $\theta$), denoting the target position of the foot
placement and the target yaw orientation of the robot's torso, respectively.
By attaching the ``heading vector" to the footstep, our method enables robot behaviours
such as lateral walking and turning in place, which would be complicated to achieve otherwise
\cite{xie2020allsteps}.
An ordered sequence of such steps is called a footstep plan. 

The footstep plans for forward walking are generated manually, simply, by
placing points alternatively to the left and right of a line segment, starting from the projection
of the robot's root on the floor and extending forward. The step length (defined by the distance
between the left and right heel in double support phase) for a sequence is specific to each robot and
must be set carefully as unrealistically small or big step length values will impede the robot from
developing a stable walking gait. The foot spread (defined by the lateral distance between the left
and right foot centers) is set to a fixed value, also specific to each robot. We draw experience from 
existing classical controllers developed for the humanoids for setting
these values: $0.15m$ foot spread and $0.35m$ step length for HRP-5P and $0.12m$ foot spread
and $0.25m$ step length for JVRC-1.
For the manually-generated straight paths, the orientation vector of each step is equal to the
``forward" direction,  i.e., the yaw orientation of the robot at which it is initialized. 

Similarly, for backwards walking, the plans can be generated by placing the points to the left
and right of a line segment extending in the backwards direction, while the orientation vector
attached to the steps remains in the forward direction. For standing in place, the trajectory
consists of only 1 step at the origin (that is $\mathbf{T_1} = \mathbf{T_2} = \mathbf{0}$). The steps
are generated manually by placing forward facing steps on the line segment at the intersection of
the frontal plane and the floor, for lateral walking plans.

Stairs are built by setting the $z$-position of the steps according to a fixed per-step height
parameter (i.e. the stair rise), decided according to a curriculum depending on the training
iteration, as explained in the next section. The stair run is fixed to be equal to the step length,
ensuring that observed target step lies exactly in the middle of the stair run.

Footstep plans for walking on a curved path are generated using existing footstep planners. We use
the footstep planner from the Humanoid Navigation ROS package \cite{hornung2012anytime} which
implements a search-based planner for 2D bipedal walking. The input to the planner is a map in
the form of a 2D occupancy grid and the initial and goal poses $(x, y, \theta)$.
A total of 1000 plans are generated by placing the start pose at the origin and randomly sampling 
the goal position from $(0, -1, -\pi/2)$ to $(0, 1, \pi/2)$ on an empty map. More complicated plans
can be generated consisting of sharper turns by placing randomly generated obstacles on the occupancy
grid map and by sampling the goal pose from a wider range.

As stated previously, the desired footstep position and desired root heading are observed by the
policy relative to the robot's root frame. When the robot successfully \textit{scores} a target
step at index $k$ of a planned sequence, the observation window slides forward by one step. So,
the step at index $k$ is replaced by $k+1$, and the policy now observes the steps $k+1$ and $k+2$
of the sequence. A step is flagged as \textit{scored}, if any foot is within the \textit{target radius}
for more than \textit{target delay} seconds. We found that a $20cm$ \textit{target radius} works for both
robots, while \textit{target delay} is set to be equal to the duration of the single-support phase to
promote synchronicity between the periodic gait and the stepping behavior. The delay is required to
allow the contacting foot to settle on the target.

\subsection{Curriculum Learning} \label{subsec:cl}
Curriculum Learning (CL) has been used successfully in multiple previous works to overcome the
problem of local minima \cite{lee2020learning, hwangbo2019learning, xie2020allsteps}, where, an
RL agent being trained for locomotion may learn to stand in place and not make any motion to avoid falling,
if presented with extremely challenging tasks in the initial stages of training.

We employ a curriculum on the sample distribution in order to learn stair climbing in a smooth manner.
Initially, the policy is only exposed to flat floor sequences - manually-generated sequences
and planner-generated curved paths. Then, after a certain number of iterations have passed, we start
to linearly increase the height of the steps from $0m$ to $\pm 0.10m$ with each iteration, in effect
building stairs (by placing ``box" type geoms at the appropriate 3D position in MuJoCo).
The iteration indices for starting and ending this linear increase in step height is determined
empirically. In our experiments, we start training on stairs after 3000 iterations and keep linearly
increasing the step height over the next 8000 iterations.
We note that there is a lot of flexibility in setting up the curriculum; the key
idea is to gradually increase the difficult of the target step sequences that the robot is made to traverse.
Mathematically, the step height at the training iteration $\textit{itr}$ is given by:

\begin{equation}
p_z = 
    \begin{cases} 
      0 & itr < 3000 \\
      k_c\cdot0.1\cdot(itr-3000/8000) & itr \geq 3000
    \end{cases}
    m,
\end{equation}

\noindent
where $k_c=\{-1, 1\}$ is a random variable that determines if the step heights are positive (for ascending)
or negative (for descending).

For simplicity, the policy is not trained for modes such as curved 3D walking,
although, we believe that it should be possible to achieve this too through an appropriate
curriculum.


\begin{figure}[t]
  \includegraphics[width=\linewidth]{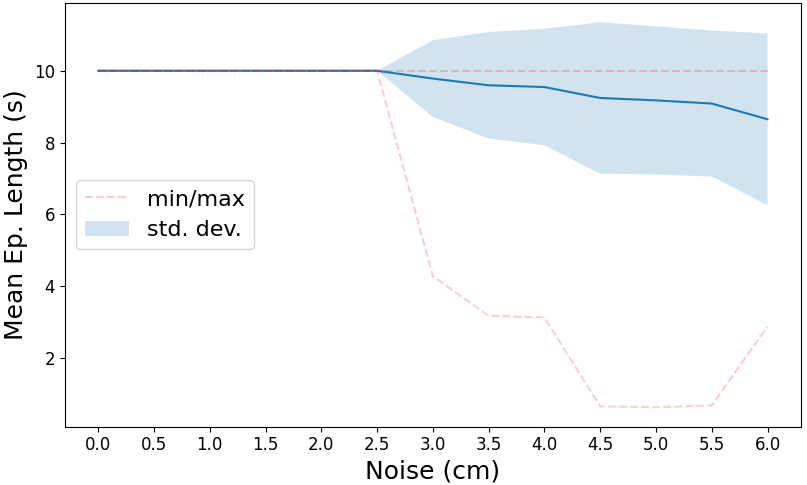}
  \caption{Robustness Test For HRP-5P with noise in height estimation. The noise is 
  drawn from a uniform distribution $\{-l, l\}$ ($l$ shown in x-axis).}
  \label{figure:terrain_noise}
  \vspace{-0.2cm}
\end{figure}


\section{Results}
\subsection{RL Policy}
\textbf{Training Details.} We used the MLP architecture to represent both the actor and critic policies,
which parameterize the policy and the value function in PPO \cite{schulman2017proximal}.
Both the MLP networks have 2 hidden layers
of size 256 each and use \textit{ReLU} activations. To limit the range of the actor's predictions, the output 
of the actor policy is passed through a \textit{TanH} layer. Each PPO rollout is of length 400 timesteps,
and each training batch holds 64 of such rollouts. The learning rate was set to 0.0001. The hyperparameters
are mostly similar to \cite{siekmann2021sim}, however, we expect that the training will be stable for
other choices of batch size and maximum trajectory length.
We also adopt the \textit{LOSS} method proposed in \cite{abdolhosseini2019learning, yu2018learning}, which 
adds an auxiliary loss term (in addition to the original PPO loss term) to enforce
symmetry by formally defining functions for obtaining mirrored states and mirrored actions.
Training the policy takes around 12 hours to collect a total of 50 million samples for learning all modes,
on a Intel Core i7-10750H CPU @ 2.60GHz with 12 cores. Simulations and optimizations are done entirely on CPU.

Our experiments show that the periodic reward composition combined with the symmetric loss term, 
is effective in generating a bipedal gait for realistic gait cycle durations.
However, backwards walking is generally more difficult to learn when combined with other modes
in a curriculum. This has been studied previously and can be overcome by methods such as DASS \cite{xie2020learning}.

\textbf{Robustness.}
We evaluate the robustness of the learned policy in two scenarios: (a) what happens when the terrain estimation
is poor? And, (b) what happens when there is noise in the robot state observation?

To answer the first question, we build a test terrain of stairs of rise $10cm$ and run $30cm$.
Then, we perform 100 rollouts of maximum episode length of 10s of a trained policy for HRP-5P, while adding 
uniformly sampled noise in the z-position of the steps.

The success rate and the average length of episodes in the tests are noted in Fig. \ref{figure:terrain_noise}
along with the range of distribution from which the noise is uniformly sampled.
A fall is detected when the root height goes below $60cm$ or when there is a self-collision between the links.
The data shows that the performance of the policy falls considerably as the noise goes beyond $\pm3cm$.
There is a higher chance of failure if the actual stair rise is higher than the target footstep than in the
case when the stair rise is lower - as the robot trips and fall due to unexpected contacts with the stair riser
in the former case.

We also simulate noise in the input signal to the policy by adding uniformly sampled noise to the 
measured joint positions and joint velocities. Trials of maximum episode length of 10s are run in the stair
climbing environment. The policy is remarkably robust here, being able to withstand noise of up to $\pm3$ degrees
before significant performance deterioration. Note that the accuracy of the motor encoders on the real hardware is much
higher than this.

\begin{figure}[t]
  \includegraphics[width=\linewidth]{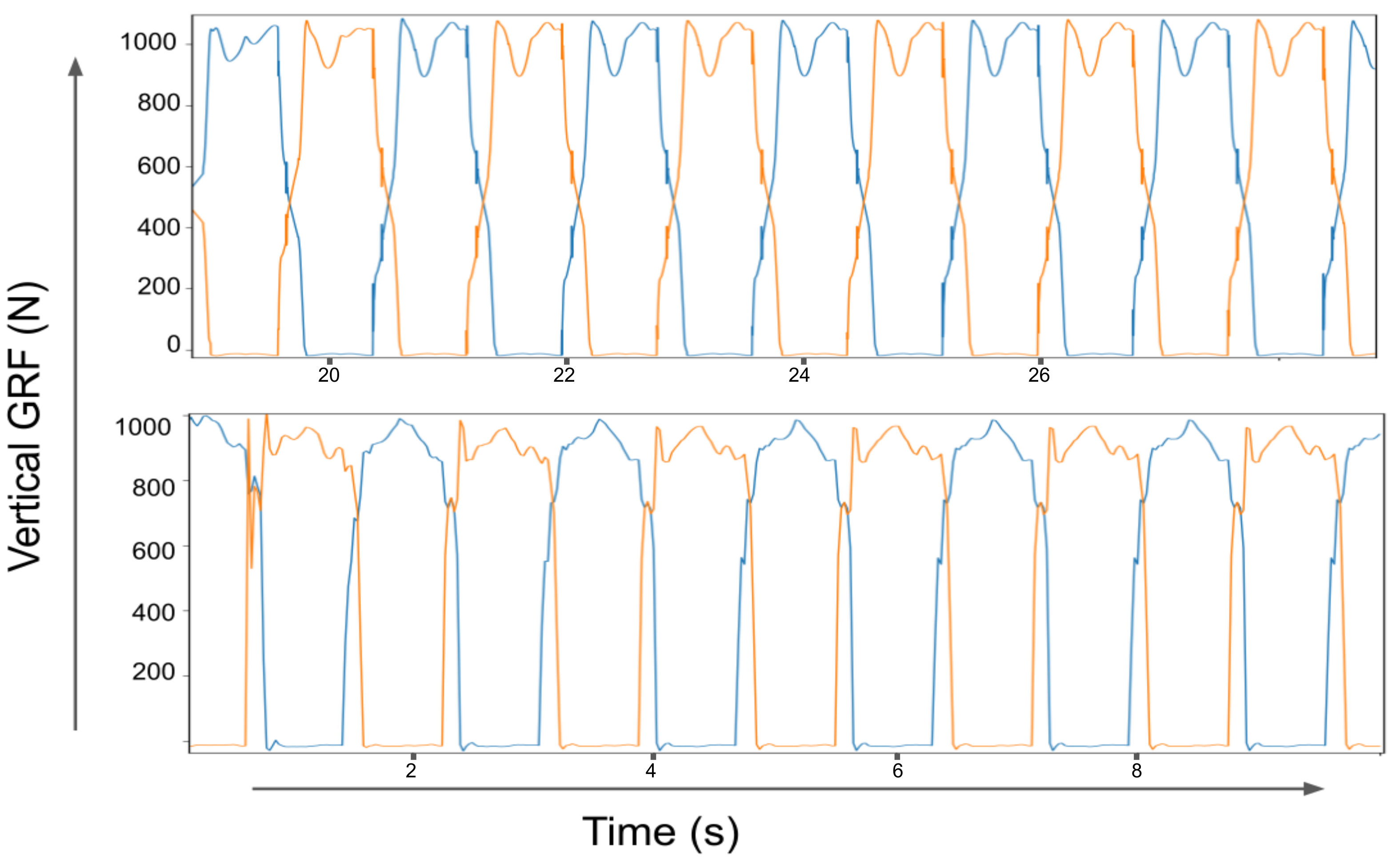}
  \caption{\textbf{Measured ground reaction forces (in simulation)} on the \textcolor{blue}{left foot} and
  \textcolor{orange}{right foot} for (a) LIPM Walking (top) and (b) RL controller (bottom).
  In both cases the peak vertical force is around 1000N (robot mass = 103kg).}
  \label{figure:feet_grf}
  \vspace{-0.4cm}
\end{figure}


\begin{table*}
\caption{Peak torques and velocities for leg joints during a 10s walk.}
\vspace{-0.4cm}
\label{table:realism}
\begin{center}
\begin{tabular}{c | c | c | c | c | c | c | c | c | c | c | c | c}
\hline
& \multicolumn{6}{c|}{Peak joint torques [N]} & \multicolumn{6}{c}{Peak joint velocity [rad/s]}\\
\hline
& Hip & Hip & Hip & Knee & Ankle & Ankle 
& Hip & Hip & Hip & Knee & Ankle & Ankle\\
& yaw & roll & pitch & pitch & pitch & roll
& yaw & roll & pitch & pitch & pitch & roll\\
\hline
Upper limit & 250 & 600 & 600 & 1420 & 300 & 250 & 10.29 & 7.85 & 8.58 & 4.66 & 8.49 & 10.19\\
\hline
Right Leg & 24.1 & 60.6 & 97.9 & 39.8 & 9.45 & 10.2 & 1.18 & 0.72 & 1.50 & 1.51 & 1.14 & 0.62\\
\hline
Left Leg & 17.1 & 55.0 & 84.4 & 76.3 & 23.9 & 11.3 & 0.37 & 0.39 & 1.57 & 2.00 & 1.25 & 0.57\\
\hline
\end{tabular}
\end{center}
\end{table*}

\textbf{Realism of Locomotion.} As the models of the robots used for training includes limits and
properties that are well-defined and approximately the same as the real robots, we expect that
the motion generated in the simulation environment will respect the limits of the physical hardware too. 
Table \ref{table:realism} shows that the torque and velocity peaks for the actuated joints are well below the 
upper limits for a 10s walk (for HRP-5P). Figure \ref{figure:feet_grf} also shows the GRF
measured by the feet force sensors in the vertical direction, for flat-terrain forward walking using,
(a) the developed RL policy and (b) the conventional approach based on the model-based LIPM (Linear Inverted Pendulum Mode)-Walking
controller \cite{caron2019stair}. The peak values in both the cases are within the safe ranges of 
the force sensors installed on the real robot's feet. 

Yet, we believe that applicability to the real system may still be a challenging task
due to several non-linear effects in the actuator dynamics (such as dry friction)
that are exceptionally difficult to model \cite{cisneros2020reliable}.

\textbf{Differences to Cassie.} Using the available model files \cite{cassiemujocosim}, we also
trained Cassie for locomotion using our framework. We found that, compared to Cassie, it is more difficult
to obtain reasonable learned motion for HRP-5P or JVRC-1, without the periodic reward terms (Eq. 
(\ref{eq:grf_reward}), (\ref{eq:spd_reward}) and the auxiliary \textit{LOSS} term for symmetry.
Further, for the humanoids, it is very important to offset the predictions from the RL policy with fixed
joint positions (corresponding to the half-sitting posture). Without this, the knee joints become stretched
and easily hit the joint limits. We also believe that bent knees are helpful for stability by providing added
compliance in the vertical direction (also the case of Cassie). The upper body reward (Eq. \ref{eq:upper_reward})
is helpful in maintaining a upright posture.

Due to a longer gait cycle period, heavier limbs and large armature values, the motion learned by
the humanoids also appears to be ``slower" than Cassie.

While Cassie has previously been trained for stair traversal without a curriculum
\cite{siekmann2021blind}, we found that CL is essential for learning a policy in our case to
achieve the desired behavior. 


\section{Conclusion}

We introduced a framework for learning an RL policy capable of
following arbitrary footstep plans along straight and curved paths. The same policy is also able
to traverse stairs, walk backwards, turn in place, and stand still. Transition between the
different modes can be achieved simply by modulating the observed target footsteps.
We demonstrate that our technique can be applied to 2 different simulated robots - HRP-5P
and JVRC-1.

%

It is important to note that our proposed approach is distinct from blind locomotion.
On the real robot, we expect the RL policy to be used in conjunction with an external footstep planner
which in turn relies on exteroceptive measurements to plan a step sequence. The footstep planner is expected
to draw out steps while considering the terrain, obstacles, start and goal positions and the
steps are fed into the closed-loop controller responsible for balancing and walking.
We believe that the methods presented in this work form a crucial milestone for
real robot deployment in the future.

\section*{Acknowledgements}
The authors thank all members of JRL for their guidance and insightful discussions. 
This work was partially supported by JST SPRING, Grant Number JPMJSP2124.


\bibliographystyle{IEEEtran}
\bibliography{IEEEabrv,bibliography.bib}
\end{document}